\documentclass{article}

\usepackage{PRIMEarxiv}

\usepackage[utf8]{inputenc} 
\usepackage[T1]{fontenc}    
\usepackage{hyperref}       
\usepackage{url}            
\usepackage{booktabs}       
\usepackage{amsfonts}       
\usepackage{nicefrac}       
\usepackage{microtype}      
\usepackage{lipsum}
\usepackage{fancyhdr}       
\usepackage{graphicx}       
\graphicspath{{media/}}     
\usepackage{algorithm}
\usepackage{algpseudocode}
\usepackage{amsmath}
\usepackage{amsthm}
\usepackage{amsmath}
\usepackage{amssymb}
\usepackage[numbers]{natbib}

\title{UniVarFL: Uniformity and Variance Regularized Federated Learning for Heterogeneous Data
\thanks{\textit{\underline{Citation}}: 
\textbf{Authors. Title. Pages.... DOI:000000/11111.}} 
}

\author{
  Sunny Gupta \\
  Koita Centre for Digital Health \\
  Indian Institute of Technology, Bombay \\
  Mumbai, Maharashtra 400076 \\
  \texttt{sunnygupta@iitb.ac.in} \\
  \And
  Nikita Jangid \\
  Department of Electrical Engineering \\
  Indian Institute of Technology, Bombay \\
  Mumbai, Maharashtra 400076 \\
  \texttt{nikitajangid@iitb.ac.in} \\
  \And
  Amit Sethi \\
  Department of Electrical Engineering \\
  Indian Institute of Technology, Bombay \\
  Mumbai, Maharashtra 400076 \\
  \texttt{asethi@iitb.ac.in} \\
}

\begin{document}
\maketitle

\begin{abstract}
Federated Learning (FL) often suffers from severe performance degradation when faced with non-IID data, largely due to local classifier bias. Traditional remedies such as global model regularization or layer freezing either incur high computational costs or struggle to adapt to feature shifts. In this work, we propose \textbf{UniVarFL}, a novel FL framework that emulates IID-like training dynamics directly at the client level, eliminating the need for global model dependency. UniVarFL leverages two complementary regularization strategies during local training: \emph{Classifier Variance Regularization}, which aligns class-wise probability distributions with those expected under IID conditions, effectively mitigating local classifier bias; and \emph{Hyperspherical Uniformity Regularization}, which encourages a uniform distribution of feature representations across the hypersphere, thereby enhancing the model's ability to generalize under diverse data distributions. Extensive experiments on multiple benchmark datasets demonstrate that UniVarFL outperforms existing methods in accuracy, highlighting its potential as a highly scalable and efficient solution for real-world FL deployments, especially in resource-constrained settings. Code: https://github.com/sunnyinAI/UniVarFL

\end{abstract}

\section{Introduction}
Federated Learning (FL) \cite{mcmahan2017communication} enables the training of deep neural networks across decentralized devices, eliminating the need for central data collection while often achieving performance comparable to centralized methods \cite{verbraeken2020survey}. However, because each device typically holds non-independent and non-identically distributed (non-IID) data, local training may converge to suboptimal local optima—a phenomenon known as \textit{client drift}—which diverges from the global objective.

To mitigate client drift, several studies have regularized local models using the global model as a reference \cite{acar2021federated, li2021model, li2020federated}. More recent work has shown that the classifier (i.e., the final layer) is particularly prone to bias in non-IID settings \cite{luo2021no}. Approaches such as using augmentation techniques \cite{yang2020gradaug} or randomly initializing and freezing classifier weights to enforce orthogonality \cite{oh2021fedbabu, he2016deep, alex2009learning} have been proposed. However, these methods generally reduce bias indirectly and often incur significant computational overhead—state-of-the-art methods like FedProx \cite{li2020federated}, SCAFFOLD \cite{karimireddy2020scaffold}, and FedDyn \cite{acar2021federated} require layer-wise weight comparisons, while MOON \cite{li2021model} and FedAlign \cite{mendieta2022local} demand additional forward passes. Moreover, many of these techniques are evaluated primarily in label-shift settings (often simulated with the Dirichlet distribution) and may not effectively address feature shifts.

To address these issues, we introduce \textbf{UniVarFL}, a novel framework that directly emulates IID-like training dynamics through two complementary regularization strategies:

\subsection{Classifier Variance Regularization}

This strategy harmonizes class-wise probability variances across clients to counteract local bias. By ensuring that the classifier’s output does not disproportionately favor a subset of classes—an effect commonly observed in non-IID settings \cite{luo2021no}. UniVarFL promotes balanced predictions. This regularization minimizes the divergence between locally optimized classifiers and the ideal IID condition, thereby reducing client drift without relying on computationally expensive global comparisons.

\subsection{Hyperspherical Uniformity}

UniVarFL also enforces a uniform distribution of feature representations on a hypersphere. This regularization discourages the encoder from overfitting to local, biased features by penalizing concentrated representations. As a result, the learned features are more evenly spread out, enhancing generalization across diverse data distributions. This approach is inspired by observations in which constrained classifier weight dynamics (e.g., via singular value decomposition) reveal degraded variance in non-IID settings \cite{oh2021fedbabu, he2016deep, alex2009learning}.

Together, these strategies enable local models to bridge the gap between IID and non-IID training conditions. Unlike prior methods that indirectly reduce bias through global regularization, UniVarFL directly promotes IID-like dynamics at the local level. Extensive evaluations on diverse benchmarks—spanning both label-shift and feature-shift non-IID scenarios—demonstrate that UniVarFL achieves state-of-the-art performance. Notably, it delivers superior accuracy, faster convergence, and enhanced computational efficiency compared to existing methods such as FedProx \cite{li2020federated}, SCAFFOLD \cite{karimireddy2020scaffold}, FedDyn \cite{acar2021federated}, MOON \cite{li2021model}, and FedAlign \cite{mendieta2022local}.

\section{Related Work}

\subsection{Federated Learning (FL)}

Federated Learning (FL) is a distributed approach that supports privacy-preserving model training by ensuring raw data remains on local devices. The canonical workflow, as illustrated by FedAvg \cite{mcmahan2017communication}, involves four main steps. Initially, a central server broadcasts the current global model to all participating clients. Each client then refines this model using its locally stored data and sends the updated parameters back to the server. Finally, the server aggregates these local updates into a new global model.

While FL encompasses a broad range of research directions—including client selection \cite{nagalapatti2022your,tang2022fedcor}, data sharing \cite{yoon2021fedmix}, privacy \cite{cheng2022differentially,karimireddy2020byzantine}, communication efficiency \cite{diao2020heterofl,wu2022smartidx,yang2021achieving}, medical applications \cite{xu2022closing}, and knowledge distillation \cite{shen2022cd2,zhang2022fine}—our work focuses on mitigating performance degradation arising from non-IID data distributions. These efforts typically involve adjustments to either the aggregation procedure or the local training process.

\subsection{Aggregation Regularization}

Aggregation-based methods maintain the standard local training routine while modifying how client models are aggregated. For instance, FedAvgM \cite{hsu2019measuring} incorporates momentum during aggregation, while FedNova \cite{yang2021achieving} normalizes local updates before averaging. In personalized FL, hypernetworks have been employed to generate client-specific parameters \cite{ma2022layer,shamsian2021personalized}. Other techniques—such as PFNM \cite{yurochkin2019bayesian} and FedMA \cite{wang2020federated} leverage Bayesian methods to align neurons across clients, while PAN \cite{li2022federated} integrates positional encoding for enhanced efficiency. Additionally, FedDF \cite{lin2020ensemble} utilizes generated unlabeled data, and CCVR \cite{luo2021no} addresses the classifier’s bias (typically the most vulnerable layer) by employing a Gaussian mixture model to calibrate artificial data.

\subsection{Local Training Regularization}

Local training regularization strategies adjust the client-side optimization without altering the aggregation step. Methods like FedProx \cite{li2020federated} add an L2 penalty between local and global model weights to limit divergence, while FedDyn \cite{acar2021federated} introduces an inner product term between consecutive local models to stabilize convergence. MOON \cite{li2021model} further refines this approach by incorporating a contrastive loss that balances representational similarities among the global model, the current local model, and the previous iteration. 

Although these techniques leverage the global model as a guiding reference, they encounter two main limitations. First, the global model---being an average of local updates---can inherit bias from its constituent models. Second, the additional computations (such as L2 distance measurements and multiple forward passes) can hinder scalability.

Recent advancements demonstrate that effective regularization need not depend on the global model. For example, FedAlign \cite{mendieta2022local} employs GradAug \cite{yang2020gradaug} for self-regularization, albeit with extra computational overhead, while FedBABU \cite{oh2021fedbabu} reduces classifier bias by randomly initializing and freezing the classifier weights. 

Inspired by these insights, our proposed framework, \textbf{UniVarFL}, operates solely at the local training level. UniVarFL directly emulates IID-like training dynamics by integrating two novel regularization strategies—\textit{Classifier Variance Regularization} and \textit{Hyperspherical Uniformity}—to mitigate non-IID degradation without incurring heavy computational costs.

\begin{figure}
    \centering
    \includegraphics[width=0.75\linewidth]{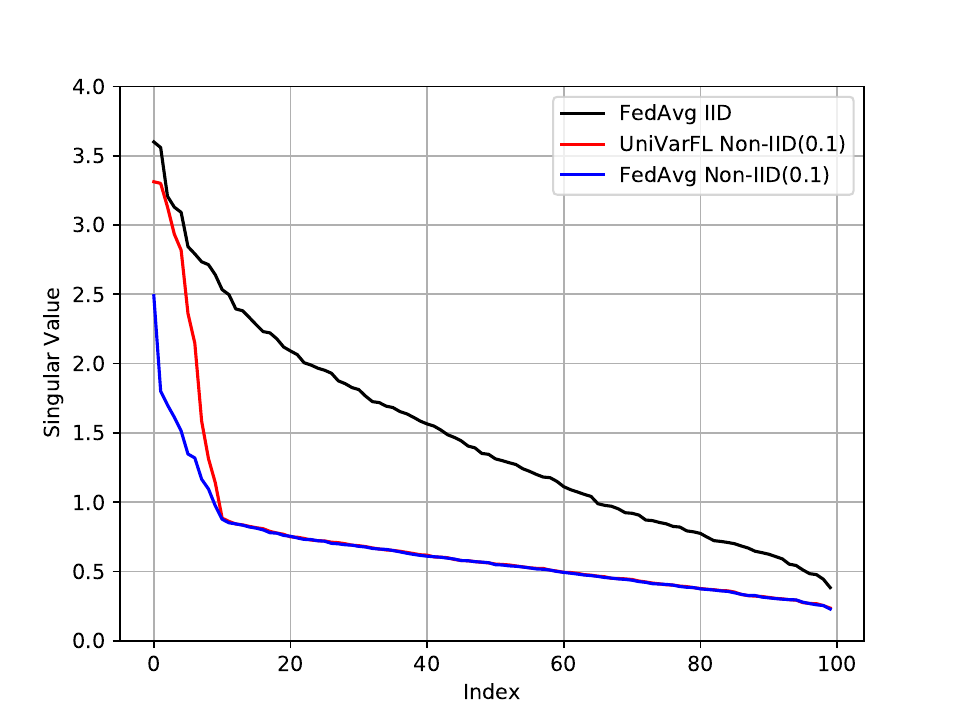}
    \caption{Presents the singular values of the classifier’s final layer weights, obtained from a model trained on the CIFAR-100 dataset.}
    \label{fig:singularplt}
\end{figure}

\begin{algorithm}
\caption{UniVarFL: Federated Learning with Variance and Hyperspherical Energy Regularization}
\label{alg:UniVarFL}
\begin{algorithmic}[1]
\Require Number of rounds $R$, number of clients $K$, learning rate $\eta$, coefficients $\mu$, $\lambda$, threshold $c$, initial global model $\theta^0$
\For{each round $r = 1$ to $R$}
    \State \textbf{Server} broadcasts global model $\theta^{r-1}$ to a subset (or all) of the $K$ clients
    \For{each client $k$ \textbf{in parallel}}
        \State Initialize local model: $\theta^r_k \gets \theta^{r-1}$
        \State Sample local dataset $\mathcal{D}_k = \{(x_i, y_i)\}$
        \For{each local training step}
            \State \textbf{1. Classifier Variance Regularization:}
            \State \quad Predict class probabilities $\hat{P}^j$ for each class $j$
            \State \quad Compute variance $\mathrm{Var}(\hat{P}^j)$
            \State \quad 
            \[
                \mathcal{L}_{\text{V}} = \frac{1}{D} \sum_{j=1}^{D} \max(0, c - \mathrm{Var}(\hat{P}^j))
            \]
            \State \textbf{2. Hyperspherical Energy Regularization:}
            \State \quad Extract features $g_\theta(x_i)$ from encoder
            \State \quad Compute angles $\theta_{i,j}$ between pairs of features
            \State \quad 
            \[
                \mathcal{L}_{\text{HE}} = \frac{1}{n^2} \sum_{i,j=1}^{n} \left(1 - \cos(\theta_{i,j}) + \epsilon\right)^{-1}
            \]
            \State \textbf{3. Total Local Loss:}
            \State \quad 
            \[
                \mathcal{L} = \mathcal{L}_{\text{CE}} + \mu \cdot \mathcal{L}_{\text{HE}} + \lambda \cdot \mathcal{L}_{\text{V}}
            \]
            \State \textbf{4. Gradient Update:}
            \State \quad $\theta^r_k \gets \theta^r_k - \eta \cdot \nabla_{\theta^r_k} \mathcal{L}$
        \EndFor
        \State \textbf{Client $k$} sends updated $\theta^r_k$ to server
    \EndFor
    \State \textbf{Server Aggregation:}
    \State \quad 
    \[
        \theta^r \gets \frac{1}{K} \sum_{k=1}^{K} \theta^r_k \quad \text{(e.g., via FedAvg)}
    \]
\EndFor
\State \Return Final global model $\theta^R$
\end{algorithmic}
\end{algorithm}

\section{Proposed Method: UniVarFL}

UniVarFL \ref{alg:UniVarFL} is designed to address two key challenges commonly encountered in Federated Learning (FL) under non-IID data distributions: (i) \emph{classifier variance degeneration}, where the classifier biases toward locally dominant classes, and (ii) \emph{feature imbalance in representations}, where encoder outputs fail to capture a uniformly distributed feature space under heterogeneous conditions. By explicitly regularizing both the classifier outputs and the underlying feature representations, UniVarFL improves accuracy, and robustness in real-world settings.

\subsection{Classifier Variance Degeneration}
In FL, model parameters (particularly those in the final layer, or ``classifier'') tend to overfit to local data distributions, resulting in skewed or degenerate weights. This phenomenon is often reflected by the rapid decay of singular values in the classifier matrix under non-IID conditions (Fig.~\ref{fig:singularplt}). To counteract this bias, we propose a \emph{Classifier Variance Regularization} term, denoted by \(L_{V}\). Intuitively, \(L_{V}\) ensures that the class-wise variance of the predicted probability distributions does not collapse, thereby preserving a broader margin of separability across classes.

Formally, let \(f_{\theta}(\cdot)\) denote the full model (encoder followed by classifier), parameterized by \(\theta\). Given a batch of inputs \(X = \{x_1, x_2, \dots, x_n\}\), the model outputs predicted class probabilities \(f_{\theta}(x_i) \in \mathbb{R}^D\) for each sample \(x_i\). 

Let \(\hat{P}^j = \{f_{\theta}(x_i)_j\}_{i=1}^n\) denote the set of predicted probabilities assigned to class \(j\) across all \(n\) samples in the batch. Then, the classifier variance regularization loss \(L_V\) is defined as:
\begin{equation}
L_V\bigl(f_{\theta}(X)\bigr) 
= \frac{1}{D} \sum_{j=1}^{D} \max\left(0,\, c - \mathrm{Var}_{x_i \sim X}(f_{\theta}(x_i)_j)\right),
\end{equation}
where \(D\) is the total number of classes, and \(\mathrm{Var}_{x_i \sim X}(f_{\theta}(x_i)_j)\) denotes the empirical variance of predicted scores for class \(j\) over the mini-batch \(X\).

The threshold \(c\) is computed from a theoretical IID reference to serve as a class-wise variance floor, anchoring predictions toward a well-calibrated distribution. Specifically, we construct a \(D \times D\) identity matrix \(A\), representing ideal one-hot class predictions, and define:
\begin{equation}
c := \frac{1}{D} \sum_{j=1}^{D} \mathrm{Var}(A_j),
\end{equation}
where \(A_j\) is the one-hot vector corresponding to class \(j\). This baseline reflects the expected variance under perfect class separability in IID settings. Incorporating this reference encourages local models to preserve diversity in predictions, mitigating overconfidence on locally dominant classes and improving robustness to data heterogeneity.

\subsection{Feature Imbalance in Representations}
While maintaining classifier variance is essential, it alone does not prevent representational collapse—especially under non-IID conditions that induce domain or feature shift. Such shifts arise when client data distributions differ in texture, background, or style, often leading to \textbf{local feature entanglement} or \textbf{mode collapse} in the encoder’s output space.

To mitigate this, we introduce a \emph{Hyperspherical Energy Regularization} term, \(L_{HE}\), which encourages encoder outputs to be angularly uniform across the hypersphere. This promotes \textbf{feature diversity} and prevents over-concentration of embeddings, thus improving generalization across heterogeneous clients.

Let \(g_{\theta}(\cdot)\) denote the encoder component of the model, mapping an input \(x_i \in X = \{x_1, \dots, x_n\}\) to its feature vector \(z_i = g_{\theta}(x_i) \in \mathbb{R}^d\), which is assumed to be \(\ell_2\)-normalized (i.e., \(\|z_i\|_2 = 1\)). The hyperspherical energy is then defined as:
\begin{equation}
L_{HE}\bigl(g_{\theta}(X)\bigr) 
= \frac{1}{n^2} \sum_{i=1}^{n} \sum_{j=1}^{n}
\left( \frac{1}{1 - z_i^\top z_j + \epsilon} \right),
\end{equation}
where \(z_i^\top z_j = \cos(\theta_{i,j})\) is the cosine similarity (inner product) between feature vectors \(z_i\) and \(z_j\), and \(\epsilon > 0\) is a small constant added for numerical stability.

By penalizing pairs of embeddings with high cosine similarity (i.e., small angular separation), \(L_{HE}\) encourages features to repel each other on the hypersphere. This geometric spreading reduces redundant local memorization and enforces distributional uniformity, which is particularly beneficial in non-IID FL scenarios where embedding diversity across clients is often lacking.

\vspace{0.5em}
\noindent \textbf{Theoretical and Empirical Justification:}
This approach is inspired by insights from hyperspherical representation learning, where angular dispersion improves generalization and robustness. Unlike contrastive losses that require negative sampling, our method achieves similar benefits in a non-contrastive, fully local manner—making it lightweight and scalable for FL settings.

\vspace{0.5em}
\noindent \textbf{Theoretical and Empirical Justification:}
This approach is inspired by insights from hyperspherical representation learning, where angular dispersion improves generalization and robustness. Unlike contrastive losses that require negative sampling, our method achieves similar benefits in a non-contrastive, fully local manner—making it lightweight and scalable for FL settings.

\subsection{UniVarFL Framework}
UniVarFL integrates these two specialized regularizers into a unified local training objective, coupling them with standard cross-entropy loss:
\begin{equation}
L = L_{CE}\bigl(f_{\theta}(X), Y\bigr) 
+ \mu \, L_{HE}\bigl(g_{\theta}(X)\bigr) 
+ \lambda \, L_{V}\bigl(f_{\theta}(X)\bigr),
\end{equation}
where:
The total loss consists of three components: \(L_{\text{CE}}\), the standard cross-entropy loss that aligns model predictions with ground truth labels; \(L_{\text{HE}}\), the hyperspherical energy term that promotes a uniform distribution of feature representations; and \(L_{\text{V}}\), the classifier variance regularization term that guards against class-wise probability collapse. The weighting coefficients \(\mu\) and \(\lambda\) control the relative importance of \(L_{\text{HE}}\) and \(L_{\text{V}}\), respectively.

During each federated round, clients minimize the above objective on their local data and transmit updated model parameters to the server for aggregation (e.g., via FedAvg). By strategically constraining both the final-layer probabilities and the intermediate feature representations, UniVarFL steers local training dynamics toward more IID-like behavior. This dual regularization approach proves especially beneficial under label-shift (where classes are imbalanced across clients) and feature-shift (where domain-specific features differ significantly among clients). Empirically, UniVarFL consistently enhances model accuracy, stabilizes convergence, and remains computationally practical for large-scale deployments—thereby offering a robust solution for modern federated learning applications.

\section{Experimental Setup}
We compare UniVarFL against established FL methods—FedAvg \cite{mcmahan2017communication}, FedProx \cite{li2020federated}, and MOON \cite{li2021model}—alongside a baseline called Freeze, which constrains the final layer in a manner similar to FedBABU \cite{oh2021fedbabu} without personalization. All experiments are conducted in PyTorch using the respective official implementations.

Following \cite{li2021model,mendieta2022local,oh2021fedbabu}, we simulate label-shift non-IID scenarios through a Dirichlet distribution. The concentration parameter $\alpha$ governs the degree of skew, with $\alpha = 0$ indicating highly non-IID data and $\alpha = \infty$ approximating IID. We select $\alpha \in \{0.01, 1.0\}$ and split each dataset 90--10 for training and validation, retaining the original validation set for testing.

\noindent
\textbf{Models.} Consistent with \cite{li2021model}, we adopt a two-layer projector (batch normalization + ReLU). For STL-10 and PACS, we use a lightweight CNN (described in Appendix~3) combined with a 256-neuron projector. For CIFAR-100 and PACS, we employ a ResNet-18 \cite{he2016deep} without its initial max-pooling layer, paired with a 512-neuron projector. All evaluations report accuracy from a single, aggregated global model.

\noindent
\textbf{Hyperparameters.} By default, STL-10 ($\alpha = 0.01$) runs on the small CNN, while CIFAR-100/PACS uses ResNet-18. Each client locally trains for $E = 10$ epochs. For STL-10, we employ 10 clients, all participating each round; we run 100 rounds of aggregation for STL-10 and 60 for CIFAR-100/PACS, guided by FedAvg validation accuracy. Reported results reflect the mean of three runs. Training uses cross-entropy loss with SGD at a 0.01 learning rate, 0.9 momentum, and $1 \times 10^{-5}$ weight decay. For regularization-based methods, additional hyperparameters are tuned on CIFAR-100: FedProx uses $\mu = 0.01$, MOON uses $\mu = 1.0$, and UniVarFL sets $\lambda = D/4$ (where $D$ is the number of classes) and $\mu = 0.5$. All experiments are performed on four RTX A6000 GPUs.

\section{Results}
\subsection*{Data Heterogeneity Analysis}
We investigate the influence of data heterogeneity on the performance of the global model under both label-shift and feature-shift scenarios. For the label-shift setting, we vary the Dirichlet concentration parameter $\alpha \in \{0.01, 1.0\}$. In contrast, feature-shift experiments utilize datasets with predefined domains (see Table~\ref{tab:tab1} for details).

In the extreme label-shift scenario ($\alpha = 0.01$), \textbf{UniVarFL} outperforms all competitors by effectively emulating an IID environment. In contrast, methods such as FedProx and MOON, which rely on global model regularization, suffer from inherited biases due to highly skewed local data. Similarly, Freeze underperforms because its fixed classifier restricts the encoder’s adaptability under severe heterogeneity.

Under a milder label-shift ($\alpha = 1.0$), regularization techniques generally perform competitively. Notably, Freeze excels on STL-10 by mitigating classifier bias in datasets with fewer classes, whereas \textbf{UniVarFL} achieves the highest performance on CIFAR-100, benefiting from its alignment-based regularization.

In feature-shift settings, conventional regularization methods provide only limited gains since they primarily address label imbalances. \textbf{UniVarFL}, by simulating IID conditions, consistently outperforms these alternatives—even those relying on biased global models—thereby demonstrating its robustness across both label-shift and feature-shift scenarios.

\begin{table*}[htbp]
    \centering
    \resizebox{\textwidth}{!}{
    \begin{tabular}{lcccccccc}
        \toprule
        \textbf{Method} & \multicolumn{2}{c}{\textbf{STL-10}} & \multicolumn{2}{c}{\textbf{CIFAR-100}} & \textbf{PACS} & \textbf{HAM10000} \\
        \cmidrule(lr){2-3} \cmidrule(lr){4-5}
        & $\alpha=0.01$ & $\alpha=1.0$ & $\alpha=0.01$ & $\alpha=1.0$ &  &  \\
        \midrule
        FedAvg   & 28.1$\pm$1.3 & 69.1$\pm$0.6 & 51.7$\pm$1.9 & 58.3$\pm$1.5 & 62.5$\pm$0.4 & 72.8$\pm$0.1 \\
        FedProx  & 26.9$\pm$0.7 & 67.6$\pm$0.8 & 50.3$\pm$0.4 & 59.2$\pm$0.9 & 58.9$\pm$0.6 & 73.5$\pm$0.3 \\
        MOON     & 26.0$\pm$0.8 & 69.4$\pm$0.4 & 47.9$\pm$1.4 & 60.0$\pm$1.0 & 63.7$\pm$1.5 & 72.9$\pm$0.2 \\
        Freeze   & 23.8$\pm$1.0 & \textbf{72.0$\pm$0.1} & 51.5$\pm$1.4 & 59.3$\pm$1.9 & 61.8$\pm$0.1 & 73.1$\pm$0.6 \\
        UniVarFL & \textbf{28.5$\pm$1.4} & 59.9$\pm$0.6 & \textbf{56.9$\pm$1.0} & \textbf{60.5$\pm$0.9} & \textbf{66.0$\pm$0.9} & \textbf{74.6$\pm$0.5} \\
        \bottomrule
    \end{tabular}
    }
    \caption{Comparison of test accuracies for $\alpha \in {0.01, 1.0}$ on the STL-10, CIFAR-100, PACS, and HAM10000 datasets.}
    \label{tab:tab1}
\end{table*}

\subsection*{Client participation}
It is common in real-world federated learning scenarios for only a fraction of the clients to participate in each aggregation round. Consequently, the central server proceeds with the aggregation process even when some clients are absent. To analyze the impact of varying client participation, we define the participation ratio $\rho$, where $\rho = 0.1$ indicates that only 10\% of the clients contribute per round. In these experiments, we employ 100 aggregation rounds for STL-10. The resulting performance is reported in Table~\ref{tab:tab2}


\begin{table}[h!]
\centering
\setlength{\tabcolsep}{5pt}
\renewcommand{\arraystretch}{1.1}
\begin{tabular}{lccc}
\toprule
\textbf{Method} & \textbf{p = 0.1} & \textbf{p = 0.5} & \textbf{p = 1.0} \\
\midrule
FedAvg   & 17.6{\small$\pm$}3.3 & 21.6{\small$\pm$}1.7 & 27.4{\small$\pm$}1.7 \\
FedProx  & 19.4{\small$\pm$}2.8 & 21.9{\small$\pm$}1.5 & 26.4{\small$\pm$}1.6 \\
MOON     & 20.6{\small$\pm$}3.5 & 23.4{\small$\pm$}1.8 & 25.9{\small$\pm$}1.4 \\
Freeze   & 21.7{\small$\pm$}3.5 & 23.6{\small$\pm$}1.7 & 23.9{\small$\pm$}1.4 \\
\textbf{UniVarFL} & \textbf{23.4{\small$\pm$}3.1} & \textbf{27.9{\small$\pm$}1.1} & \textbf{28.5{\small$\pm$}1.4} \\
\bottomrule
\end{tabular}
\vspace{1mm}
\caption{Test accuracy (\%) on STL-10 under different client participation rates ($\alpha = 0.01$).}
\label{tab:tab2}
\end{table}

\section{Theoretical Analysis under Nonconvex Objectives}
\label{sec:theory}

In this section, we provide a theoretical analysis of \textsc{UniVarFL} under a more realistic \emph{nonconvex} setting.  Specifically, we show that when each local objective $L_k(\theta)$ is \emph{smooth} and when certain assumptions on local updates and data heterogeneity hold, \textsc{UniVarFL} converges to a \emph{stationary point} of the global objective at a rate $\mathcal{O}(1/\sqrt{T})$.

Recall that client $k$ holds local data $\mathcal{D}_k$ and defines its local objective as:
\[
L_k(\theta) = L_{\text{CE}}(f_\theta(X_k), Y_k) + \mu L_{\text{HUR}}(g_\theta(X_k)) + \lambda L_{\text{CVR}}(f_\theta(X_k)),
\]
where $L_{\text{CE}}$ is the standard cross-entropy loss, $L_{\text{HUR}}$ denotes the hyperspherical uniformity regularization, and $L_{\text{CVR}}$ is the classifier variance regularization term. The global objective is a weighted average of local losses:
\[
L(\theta) = \sum_{k=1}^K \frac{n_k}{N} L_k(\theta), \quad \text{where } N = \sum_{k=1}^K n_k.
\]

In practical federated learning scenarios—particularly with deep neural networks—the objective $L(\theta)$ is typically \emph{nonconvex}. Therefore, we aim to establish convergence to a \emph{first-order stationary point}. To do so, we adopt the following standard assumptions:

\vspace{0.5em}
\noindent \textbf{Assumptions.}
\begin{itemize}
    \item \textbf{(A1) $L$-Smoothness of Local Objectives.} Each local loss function $L_k(\theta)$ is $L$-smooth, i.e., for all $\theta_1, \theta_2$,
    \[
    \|\nabla L_k(\theta_1) - \nabla L_k(\theta_2)\| \leq L \|\theta_1 - \theta_2\|.
    \]
    This assumption ensures well-behaved gradients and is commonly satisfied in deep models through design choices like weight normalization, gradient clipping, and smooth activation functions.

    \item \textbf{(A2) Bounded Local Gradient Variance.} Let $\theta_{k,\tau}$ denote the model after $\tau$ local steps on client $k$. Then, for some constant $\sigma^2 > 0$,
    \[
    \mathbb{E} \bigl[ \|\nabla L_k(\theta_{k,\tau}) - \nabla L_k(\theta_k^\star)\|^2 \bigr] \leq \sigma^2,
    \]
    where $\theta_k^\star$ is the local optimum for client $k$. This assumption bounds the stochasticity introduced by local updates, enabling stable global aggregation.

    \item \textbf{(A3) Bounded Gradient Heterogeneity.} Let $\theta^\star$ denote a global reference point (e.g., a stationary point of $L(\theta)$), and define the average gradient as $G^\star = \frac{1}{K} \sum_{k=1}^K \nabla L_k(\theta^\star)$. Then,
    \[
    \frac{1}{K} \sum_{k=1}^K \left\| \nabla L_k(\theta^\star) - G^\star \right\|^2 \leq \zeta^2,
    \]
    for some constant $\zeta^2 > 0$. This captures the inter-client heterogeneity and ensures that local gradients do not diverge excessively from the global trend.
\end{itemize}

\noindent These assumptions follow precedents in the federated optimization literature~\citep{karimireddy2020scaffold, li2020federated} and allow us to analyze the convergence behavior of our method.

\paragraph{Local Updates.}
During training, each client performs $E$ local SGD steps with learning rate $\eta$, producing an updated local model $\theta_k^{(t,E)}$ starting from the global model $\theta^{(t)}$. The server then aggregates the updated models using weighted averaging:
\[
\theta^{(t+1)} = \sum_{k=1}^K \frac{n_k}{N} \, \theta_k^{(t,E)}.
\]

\subsection{Nonconvex Convergence: Main Result}
We analyze \textsc{UniVarFL} by tracking the evolution of the global objective $L(\theta^{(t)})$ across communication rounds. Under assumptions (A1)–(A3), we establish the following convergence guarantee:

\newtheorem{theorem}{Theorem}
\begin{theorem}[Convergence to Stationary Points]
\label{thm:nonconvex}
Suppose each $L_k(\theta)$ is $L$‐smooth, and let \textsc{UniVarFL} run for $T$ rounds, each with $E$ local SGD steps of learning rate $\eta$.  Under the bounded variance and heterogeneity assumptions, if $\eta$ and $E$ are chosen so that $\eta\,E\,L$ is not too large (e.g., $\eta\,E\,L \le c$ for some small constant $c$), then there exists a constant $C$ (dependent on $L,\sigma,\zeta$) such that
\[
  \min_{0\le t < T} 
  \mathbb{E}\bigl[\|\nabla L(\theta^{(t)})\|^2\bigr]
  \;\le\;
  \frac{C}{\sqrt{TE}}.
\]
\end{theorem}

In other words, the \emph{expected gradient norm} of the global model converges to $0$ at a rate of $\mathcal{O}(1/\sqrt{T E})$, implying that after sufficient rounds, \textsc{UniVarFL} finds an approximate stationary point of $L(\theta)$.  This result is consistent with classical nonconvex SGD analysis and existing work on local/parallel SGD.

\section{Conclusion}
In this paper, we introduced \textbf{UniVarFL}, a new framework for mitigating non-IID challenges in Federated Learning by directly emulating IID-like training dynamics at the local level. Unlike prior approaches that rely on explicit global models or freeze specific layers, UniVarFL promotes class-wise probability variance through \emph{classifier variance regularization} and encourages a more balanced feature space via \emph{hyperspherical uniformity}. By jointly addressing classifier bias and feature imbalance, our method not only achieves state-of-the-art performance under both label-shift and feature-shift conditions but also maintains low computational overhead. Extensive experiments confirm that UniVarFL offers sizable gains in accuracy and convergence speed while remaining highly scalable, making it a promising solution for real-world, resource-constrained FL applications.

\nocite{langley00}



\bibliographystyle{unsrt}  
\bibliography{references}

\end{document}